\documentclass[journal,comsoc]{IEEEtran}
\usepackage[T1]{fontenc}
\usepackage{amsmath}
\usepackage[cmintegrals]{newtxmath}
\usepackage{url}
\usepackage{soul,color,graphicx,float,epstopdf}
\usepackage{tablefootnote,textcomp,gensymb}
\usepackage{eurosym,booktabs,multirow}
\usepackage[caption=false]{subfig}
\usepackage{amsfonts} 
\usepackage{algorithmic}
\usepackage{array}
\usepackage{stfloats}
\usepackage{float}
\usepackage{mathtools}
\usepackage{graphicx}
\usepackage{pdfpages}
\usepackage{subfig} 
\usepackage{comment}
\usepackage{balance}
\usepackage{xcolor}
\usepackage{comment}
\usepackage[normalem]{ulem}
\usepackage[implicit=false]{hyperref}

\usepackage[absolute, overlay]{textpos}
\usepackage{tikz}
\textblockorigin{\paperwidth}{0.0 pt}

\usepackage{tikz}
\usetikzlibrary{arrows,shapes,positioning,shadows,trees}

\tikzset{
  basic/.style  = {draw, text width=4cm, font=\sffamily, rectangle},
  root/.style   = {basic, rounded corners=2pt, thin, align=center,
                   fill=gray!20},
  level 2/.style = {basic, rounded corners=6pt, thin,align=center, fill=blue!10,
                   text width=9em},
  level 3/.style = {basic, thin, align=left, fill=yellow!5, text width=10em}
}

\definecolor{Orange}{rgb}{1,0.5,0}
\newcommand{\todo}[1]{\textsf{\textbf{\textcolor{Orange}{[#1]}}}}

\newcommand{\PZ}[1]{\textcolor{blue}{#1}}

\usepackage{eso-pic}

\begin{document}

\title{Sim2real for Reinforcement Learning Driven Next Generation Networks}

\author{Peizheng~Li,~\IEEEmembership{}
        Jonathan~Thomas,~\IEEEmembership{}
        Xiaoyang~Wang,~\IEEEmembership{}
        Hakan~Erdol,~\IEEEmembership{} 
        Abdelrahim~Ahmad,~\IEEEmembership{}\\
        Rui~Inacio,\IEEEmembership{}
        Shipra~Kapoor,~\IEEEmembership{}
        Arjun~Parekh,~\IEEEmembership{}
        Angela~Doufexi,~\IEEEmembership{}
        Arman~Shojaeifard,~\IEEEmembership{}
        and~Robert~Piechocki ~\IEEEmembership{}
        
\thanks{Peizheng~Li, Jonathan~Thomas, Xiaoyang Wang, Hakan~Erdol, Angela~Doufexi and Robert~Piechocki are with the Department of Electrical and Electronic Engineering, University of Bristol, UK (e-mail: \{peizheng.li, jonathan.david.thomas, xiaoyang.wang, hakan.erdol, A.Doufexi,  R.J.Piechocki\}@bristol.ac.uk).

Abdelrahim~Ahmad and Rui~Inacio are with Vilicom UK Ltd. (email: \{Abdelrahim.Ahmad, Rui.Inacio\}@vilicom.com).

Shipra~Kapoor and Arjun~Parekh are with Applied Research, BT, UK (email: \{shipra.kapoor, arjun.parekh\}@bt.com).

Arman Shojaeifard is with InterDigital Communications, Inc. (email: arman.shojaeifard@interdigital.com).
}}




\maketitle
\begin{tikzpicture}[remember picture, overlay]
  \node[minimum
  width=4in,fill=white!100,text=black,font=\normalsize, align=left] at ([yshift=-1cm, xshift=-2cm]current page.north)  {This work has been submitted to the IEEE for possible publication.\\
  Copyright may be transferred without notice, after which this version may no longer be accessible.};
\end{tikzpicture}
\begin{abstract}
The next generation of networks will actively embrace artificial intelligence (AI) and machine learning (ML) technologies for automation networks and optimal network operation strategies. The emerging network structure represented by Open RAN (O-RAN) conforms to this trend, and the radio intelligent controller (RIC) at the centre of its specification serves as an ML applications host. Various ML models, especially Reinforcement Learning (RL) models, are regarded as the key to solving RAN-related multi-objective optimization problems. However, it should be recognized that most of the current RL successes are confined to abstract and simplified simulation environments, which may not directly translate to high performance in complex real environments. One of the main reasons is the modelling gap between the simulation and the real environment, which could make the RL agent trained by simulation ill-equipped for the real environment. This issue is termed as the \emph{sim2real} gap. This article brings to the fore the sim2real challenge within the context of O-RAN. Specifically, it emphasizes the characteristics, and benefits that the digital twins (DT) could have as a place for model development and verification. Several use cases are presented to exemplify and demonstrate failure modes of the simulations trained RL model in real environments. The effectiveness of DT in assisting the development of RL algorithms is discussed. Then the current state of the art learning-based methods commonly used to overcome the sim2real challenge are presented. Finally, the development and deployment concerns for the RL applications realisation in O-RAN are discussed from the view of the potential issues like data interaction, environment bottlenecks, and algorithm design.
\end{abstract}


\section{Introduction}
\label{sec:introduction}
The emergence of the 5G network and beyond will undoubtedly have a huge impact on future digital infrastructures. As a part of network architecture evolution, the radio access network (RAN) is expected to transfer from proprietary networks to flexible, agile and open-source networks.  One representative example of embodying such changes is Open RAN (O-RAN). As the proposal for years of industry group exploration, O-RAN defines additional interfaces based on 3GPP new radio (NR) specifications, focusing on a functional split called 7.2x to achieve efficiency and performance dividends. The base station (BS) functionalities are virtualized as network functions and divided across multiple network nodes: central unit (CU), distributed unit (DU) and radio unit (RU) \cite{bonati2021intelligence}.
O-RAN embraces artificial intelligence/machine learning (AI/ML) at the very beginning of the standardisation. It introduces intelligence and automation for 5G and beyond by leveraging data collection capabilities from standardized open interfaces. Two types of radio intelligence controllers (RIC) are brought into the hierarchical architecture which serves as the host of ML models in the form of microservice applications called rApps and xApps. The two RICs are designed to meet the application requirements of different time sensitivity. 

Simulation results of some state of the art research demonstrate the superiority of ML over traditional methods. Different model designs adopting supervised learning (SL), unsupervised learning (UL) and reinforcement learning (RL) paradigms are considered.  For example, resource management, association control, adaptive slicing in the network layer, modulator, demodulator and beamforming design in the physical layer. SL is used for prediction or classification tasks from labelled datasets; UL is used to learn hidden patterns from unlabeled data, and RL learns optimal decision-making policies to maximize long-term rewards.

However, for the RL applications, most successes are confined to simulated environments rather than real-world settings. Current studies rely on abstracted environments by simplifying the modelling of each component in the involved tasks, to train the corresponding RL agent. This inevitably limits the generalisation of the RL models. In other words, the model trained in a simulation environment would be unfit to deploy in the real environment, which is also termed the sim2real issue. Whereas the Sim2real gap affects all ML models, it is particularly acute for RL. This is principally down to the active sequential decision-making role of RL. Even the smallest data distribution shifts between simulations and reality might lead to catastrophic decisions with a run-away effect. 
This issue has been acknowledged by various communities working with RL models. Arguably, the most representative example is a visual-based robot arm control. The inconsistent modelling of the angle of camera view, lighting, and materials of the bot arm between simulation and real environment would cause the failure of the trained control policy. Such failure mode should be expected when RL agents are used in wireless communications and O-RAN. However, there has not been much discussion of the sim2real issue thus far.
Furthermore, it is worth noting that the RL-based applications also introduce various challenges for the entire RAN network operations, including but not limited to the efficient training, reliable validation, periodic management and orchestration of ML models, which is also part of the macro sim2real gap. Therefore, this article focuses on analyzing the sim2real problem of RL in the context of O-RAN. The contributions of this article are summarised below:
\begin{itemize}
    \item This article reviews RL driven applications for the next generation networks.
   \item  We highlight the role of high-fidelity digital twins (DT) as a key part of the solution for sim2real, and elaborate in its design principles. 
   \item A case study for sim2real caused by user equipment (UE) distribution shift is demonstrated. Meanwhile, a representative case of DT-assisted training with the sim2real element is discussed.
    \item A review of the latest learning-based methodologies to overcome the sim2real challenge is presented. 
    \item Potential issues of RL applications in the real networks are discussed from the perspective of data collection, environment and algorithm design.
\end{itemize}


 \begin{figure*}[t]
    \centering
    \includegraphics[width=1\textwidth]{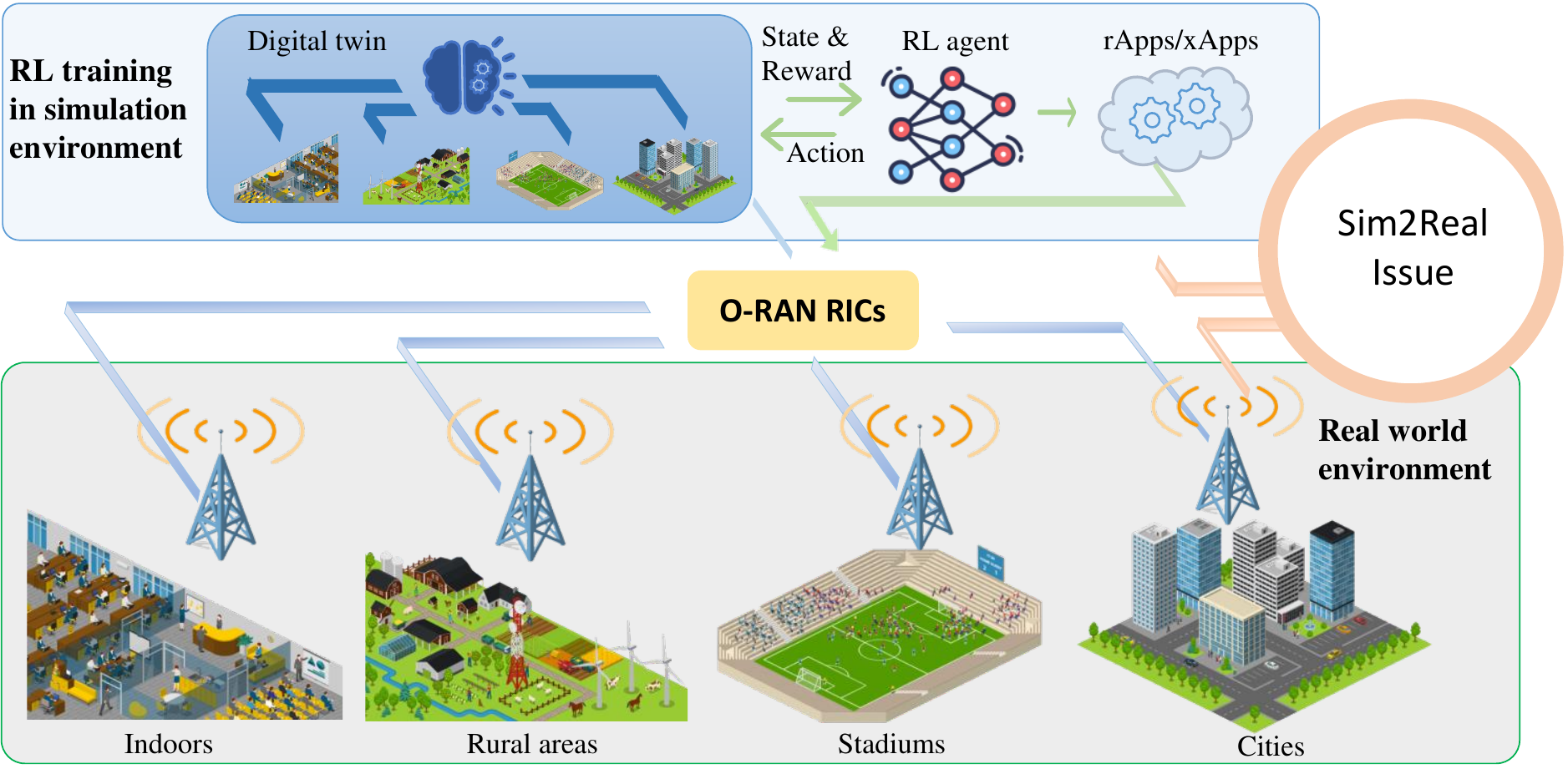}
    \caption{Next-generation of cellular networks actively embraces the ML, but the ML applications developed in the simulation environment are usually unfit for the real environment, which brings the sim2real concern.}
    \label{fig:sim2real_demo}
\end{figure*}

\section{RL driven cellular networks}
While the amount of wireless traffic grows exponentially, the variety of service demands drastically increases as well by integrating various wireless-enabled devices such as smartphones, drones, wearable electronics, sensors, etc. A variety of service schemes occurs in emerging RAN technologies to fulfil such demands. RL-based approaches are believed to be effective solutions for intelligent RAN management \cite{Chen2019ML_for_WN}.

O-RAN ecosystem contains a comprehensive control mechanism which can use various decision-making algorithms to improve QoS and QoE. Therefore, there is a number of applications for traffic management from radio resource management to RAN network slicing in O-RAN architecture. RL-based approaches are mostly pre-trained in an offline environment before deployment. If the environment is well-designed, the performance of RL agents potentially outperforms heuristic or rule-based algorithms, especially if the problem requires broad information about the state of the network during decision making \cite{Tanver2022RL_for_HO}.   

\subsection{RL Applications in Next Generation Cellular Networks}
The introduction of RL in the next-generation cellular networks will bring profound changes in every stack, namely, a shift from being static and stiff to a data-driven, dynamically sensing and self-optimising network. 
The application of RL models will undoubtedly contribute to the realization of autonomous networks.

The state of the art communication system embodies features of hierarchical and self-contained functions, like encoding, modulation,  transmission, demodulation, decoding, corresponding channel measurement etc. Each function is modelled with well-defined mathematical formulas, which achieves its local optimum and approaches the Shannon limit. However, the global optimization for the multiple-functions combined objective is difficult to realise.  One of the reasons is that such optimization problems are largely NP-Hard problems, which are difficult to solve analytically \cite{li2021rlops}.

For the 5G and beyond, such high-standard communication protocols will put forward an urgent need for similar global optimization and dynamic adjustment. It can be expected that the RL-based algorithms will shape the self-organizing networks (SON) with dynamic resource al
location properties like agile slicing, dynamic spectrum sensing, random access channel, and load balancing optimization. Furthermore, besides these use cases, there are new problems with the advancement of 5G networks i.e. dual-connectivity, and traffic steering. While similar problems were studied within former radio access technologies (RAT), such problems in 5G and beyond will become significantly more sophisticated on larger scales.

For example, handover operation in legacy RATs is triggered mostly based on UEs' channel quality indicators (CQI). Algorithms mostly focused on preventing the ping-pong effect while triggering handover operation. Nevertheless, the dual connectivity feature of 5G enables UEs to attach to one cell for mobile broadband service and attach to another cell for voice applications simultaneously. Besides these, if cells have massive-MIMO and limited resources, the handover algorithm should process a number of factors before getting triggered. RL agents are capable of collecting extensive information about network state and producing an action that will cause optimal or near-optimal outcomes for the UEs. Therefore, RL driven solutions can potentially provide a better solution for the overall scenario \cite{Tanver2022RL_for_HO}.

\subsection{Training RL Agents}
\label{subsec:Productionising RL}
Training RL agents are data-intensive and typically requires millions of interactions with a target environment. 
This sample in-efficiency coupled with safety concerns associated with agent exploration present significant difficulties for training in the real world. 
This is further exacerbated by the requirement for optimization of hyper-parameters like the algorithm, learning rates, neural network architecture and many others.
Finding a near-optimal/optimal hyper-parameter configuration is generally accomplished through search methodologies like grid-search or Bayesian optimisation. 
This further increases our requirement for environment interactions and leaves us in a situation where \textit{tabula-rasa} deployment in the real world is likely infeasible with current methodologies. 
As such, training within simulated environments is typically an essential part of any application development pipeline. 

By incorporating a simulator into our development pipeline, we can enjoy a number of benefits which are listed below. 
\begin{enumerate}
    \item \textbf{Exploration:} RL algorithms require exploration in order to learn about their environment. Exploration, by definition, is risky. It requires the execution of actions that have potentially unknown outcomes and could be unrecoverable. In a simulated environment, we can reset the environment. 
    \item \textbf{Parallelization:} Sample efficiency is a crucial concern within RL, where agents typically take considerable time to train. Simulation supports the utilization of several environments in parallel operating at increased clock rates can reduce the wall-clock time required for training.
    \item \textbf{Validation:} Demonstrating and proving RL application capability in a range of scenarios is likely to be an essential part of the development cycle. Simulation capability allows us to accomplish this in a range of scenarios including those that are potentially rare. Through this type of process, we can quantify and potentially mitigate risks associated with RL applications which will be important for all network stakeholders.  
\end{enumerate}

\subsection{Digital Twins}
An idea that has been pervasive throughout the telecommunications and wider industry communities is that of DT. A DT is defined as a digital representation of a physical item or assembly using integrated simulations and service data \cite{erkoyuncu2018digital}. The standardization of wireless network DT is in progress, but its design should basically conform to the following principles:
\begin{itemize}
      \item \textbf{Extendibility}: DT maintains open interfaces for different surrogate models and functions to quickly and flexibly construct a multi-functional platform to verify the deployment of external models.
    \item \textbf{Interactivity}: Being used as a simulation platform, DT is the training venue for ML models and can respond appropriately to RL agents with feasible actions. 
    \item \textbf{What-if}: It reflects the ability to see and analyse what would have happened if a different action had been taken or different configurations of the network had occurred, which enables the adaptive modelling of potential issues of radio traffic, network security, energy efficiency and other new disruptive services.
     \item \textbf{Scalability}: The DT should be extended to the large scale networks without loss of fidelity and simulation speed. Hence, hardware acceleration should be a design option.
    \item \textbf{Continuous refinement}: DT should be able to receive real-world data and constantly self-correct the accuracy of the surrogate models.
\end{itemize}
So, DTs are a natural extension of simulators for RL development. They provide an environment which should both approximate and is coupled with the real network state. Through this, we may hope to obtain an RL agent which is better suited to the network than may be achievable through a bespoke simulation engine. Additionally, the inclusion of a DT into the application pipeline supports model validation and continuous testing in what is likely to be a diverse ecosystem of interacting and changing applications.

\section{Sim2real}
\subsection{What is Sim2real?}
Within the wider RL community, training within simulated environments is generally considered a necessary step to develop effective RL agents, and this is for largely similar reasons\footnote{Additional reasons may include equipment degradation throughout training runs.} to those indicated in Section \ref{subsec:Productionising RL}.
Naturally, this raises a number of questions concerning the quality of simulators which we have listed below: 
\begin{enumerate}
    \item \textbf{Quantification:} How may we quantify the quality of the simulator's approximation of the real environment? What is the \textit{Sim2Real} gap?
    \item \textbf{Minimization:} Can we minimise the difference between simulation and the real environment?
    \item \textbf{Transferability:} Are the policies transferable? How do our simulator-trained agents perform in the real world? 
\end{enumerate}
These questions are fundamentally what \textit{Sim2Real} considers. 
Telecommunications is no different to any other application domain, in which these questions and challenges persist. Their importance may even be elevated, especially when considering communications networks as critical national infrastructure. Figure \ref{fig:sim2real_demo} illustrates the impact of sim2real in the next generation of access networks.

\subsection{Case Study 1: Sim2Real in an O-RAN BBU Pooling}

In this section, we present an example of the sim2real gap and its negative effect on the RL-based RIC in O-RAN. As mentioned in Section~\ref{sec:introduction}, O-RAN facilitates software-defined, AI-driven RIC through additional open interfaces, in which RL could be applied to develop dynamic control policies via learning. Baseband Units (BBU) pooling is a long-existing problem in cloudified networks, where cloudified BBUs are deployed on a common hardware pool. O-RAN enables flexible mapping from Radio Units (RUs) to BBUs to achieve RAN elasticity, bringing benefits including higher utilization, lower power consumption, lower overall cost and better resiliency~\cite{oranwp1}.

As a case study, an RL-based algorithm has been developed for BBU pooling management in O-RAN. We firstly build a simulated environment with a fixed set of RUs and dynamic UE distributions. UE distributions are extracted from historical network data to form changing traffic patterns. An RL agent is trained in this simulated environment through interactions. After training, the model would be deployed to perform BBU pooling without further tuning or exploration.

\begin{figure}[ht]
    \centering
    \includegraphics[width=2.8in]{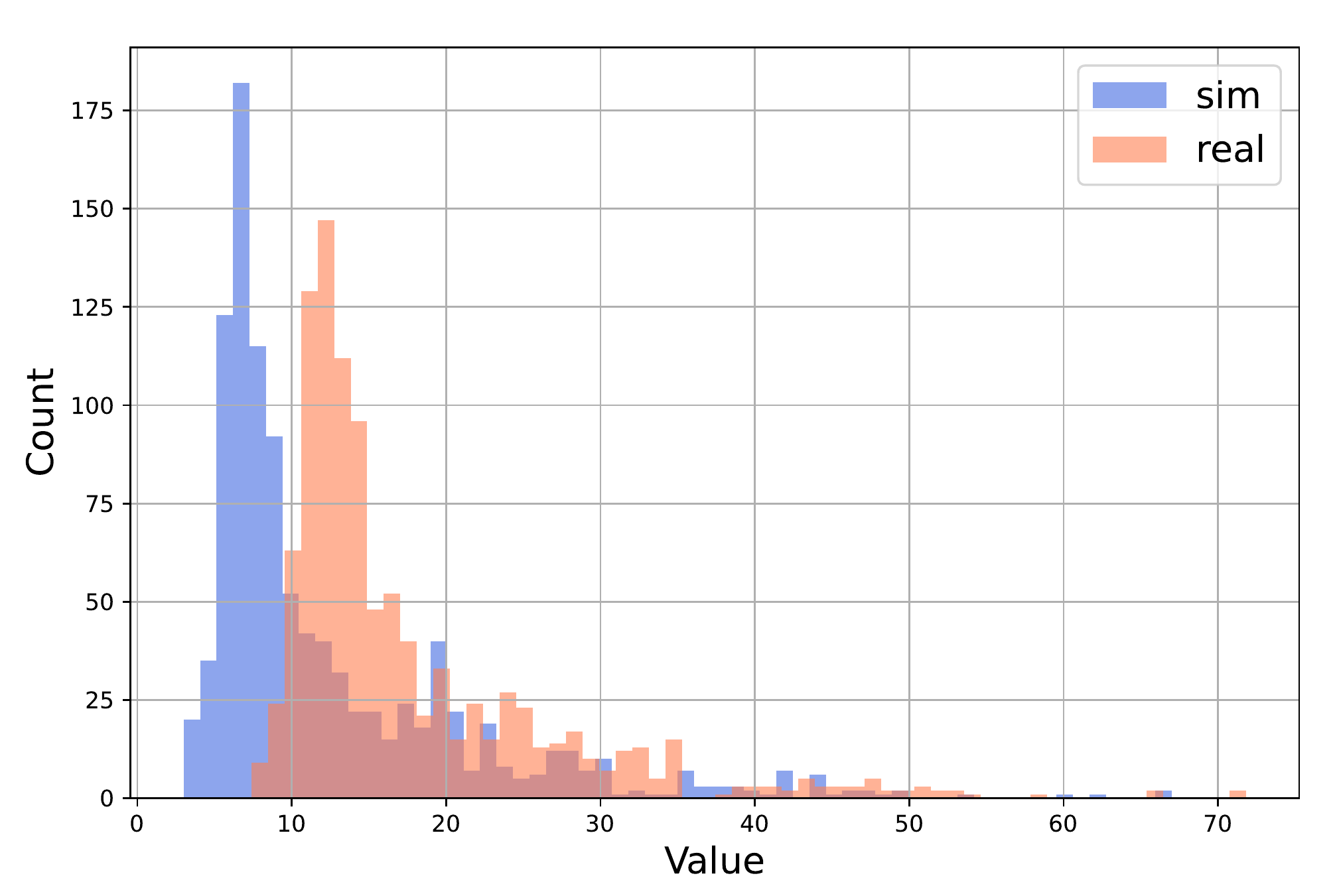}
    \caption{Histograms of traffic in the simulated environment and the ``real'' environment.}
    \label{fig:ue_dist_sim2real}
\end{figure}

\begin{figure}[ht]
    \centering
    \includegraphics[width=2.8in]{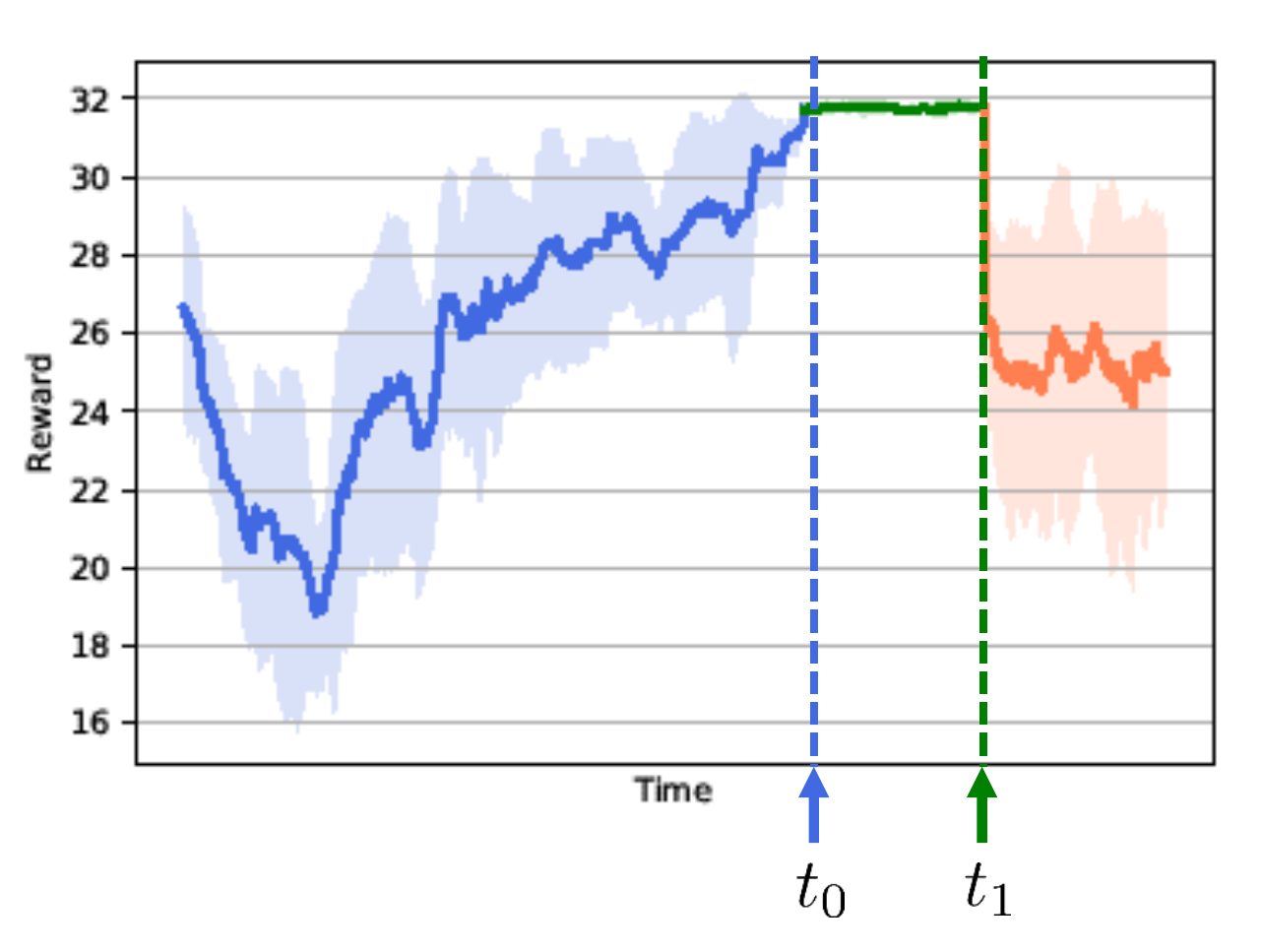}
    \caption{Reward of the RL agent for BBU pooling. Training in the simulated environment terminates at $t_0$; From $t_0$ to $t_1$, the trained model is validated in the simulated environment. After $t_1$, the UE distribution is modified, showing the results for deploying the trained RL agent to the real environment.}
    \label{fig:sim2real_reward}
\end{figure}

To mimic the variation of traffic in live networks, we add Gaussian additive noise to the simulated environment, formulating the change of UE distributions in a ``real'' scenario. Figure~\ref{fig:ue_dist_sim2real} shows the histogram of traffic in the simulated environment and the ``real'' environment, where a distribution shift exists. Figure~\ref{fig:sim2real_reward} shows the reward during training in the simulated environment, the reward when deploying the trained model to the simulated environment and to the ``real'' environment. The drop in performance is observed when the UE distribution changes. Without further training, the performance of the RL agent in the ``real'' environment will not be comparable with that in the simulated environment. In this case, the simulator provides an approximation to the real environment, where the difference is determined by the noise level. We show that the policy learned in the simulated environment is not directly transferable to the real environment.

\subsection{Case Study 2: Sim2Real in Magnetic Control of a Tokamak plasma \cite{Degrave2022}}
A particularly compelling example of the design and deployment of an RL agent for a complex control task is discussed within \cite{Degrave2022}. They consider the challenge of magnetic confinement of a high-temperature plasma within a tokamak to enable further pursuit of nuclear fusion. Of significant relevance to our manuscript is the discussion of Sim2real and how they handle the gap. Their approach involves multiple methods to overcome various challenges which originate from their discrepancies. These include (1) unstable parameter configurations within their test-bed, (2) training methodologies to overcome specific parameters which aren't known a-priori in the real world and (3) considerations for model inference time which is restricted by operational requirements. 

Firstly, for unstable parameter configurations, they discuss a method called 'learned-region avoidance' which encourages their agent through reward-shaping to learn a policy within simulation which avoids a region which is unstable and difficult to model. It is worth noting that this was only discovered through iterative training and deployment cycles, which likely attests to a requirement for specific deployment methodologies for RL agents. 

Secondly, there are certain latent parameters which the authors were unable to know in advance, as such they utilised domain randomisation. Through this, they are able to create a model which was robust to their variation.

Finally, there are hardware asymmetries between the simulated and real system which further complicate development. Most significantly, their RL agent has to be able to produce action within 50$\mu$s on the real system. This significantly restricts the size of their policy network which they partially alleviate through utilising a significantly larger LSTM-based critic which is able to handle non-markovian environment dynamics and isn't required at deployment time. Although, beyond the requirements for O-RAN real-time applications, this still may represent a key concern especially when considering computational resource contention.

 \begin{figure*}[t]
    \centering
    \includegraphics[width=1\textwidth]{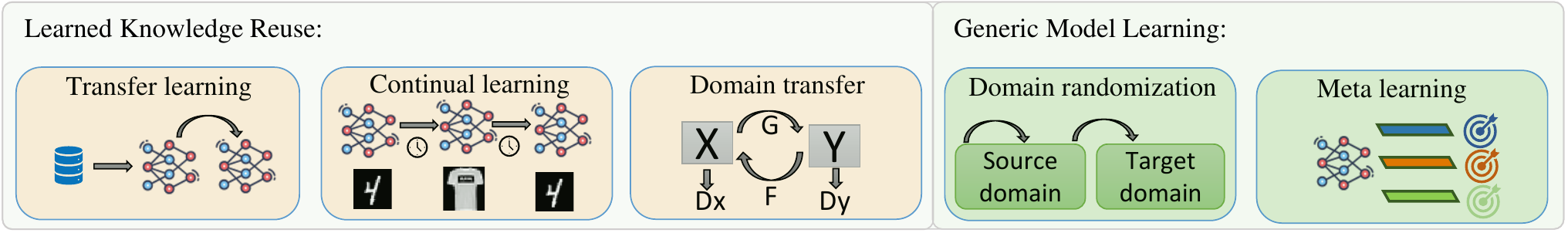}
    \caption{Demonstrations of different types of learning methods can be used to deal with the sim2real issue.}
    \label{fig:learning methods}
\end{figure*}
\section{Learning-based Methods for Sim2real}
\label{sec:learning-based methods}
The most straightforward way to achieve zero-shot or direct transfer without re-training is to build a realistic DT. But considering the multiple uncertainties of cellular network modelling, this expectation is difficult to achieve. So, the solution should be found from the perspective of the learning methods. A common approach is to reuse previously learned knowledge and transfer it to new tasks to achieve better performance on new tasks. Another approach is to learn as much as possible a general model from a set of tasks. Then for model deployment from simulation to reality, only few-shots are needed to re-adapt the model to meet the task requirements. It is worth noting that such learning methods always overlap with each other, and there are no strict classification boundaries, which means that the above two categories are often used in conjunction with each other. Figure \ref{fig:learning methods} depicts the functional and structural differences of these learning methods that are hopefully to be used to deal with the sim2real issues. More technical discussions are given below.
\subsection{Reusing Knowledge from Learned Models}
\subsubsection{Transfer Learning Related Methods}
The basic mechanism of transfer learning is to transfer the knowledge/skills from source domains to a different but relevant target domain to improve the performance of target learners and reduce the data dependency of target learners. The simplest form is to fine-tune the model by adjusting a part of the neural network parameters, or by adding new layers and freezing preceding layers. One technique widely used in transfer learning is the (supervised) domain adaptation, which refers to adapting one or more source domains to transfer knowledge and improve the performance of the target learner \cite{zhuang2020comprehensive}. Another option is knowledge distillation, a particular case of transfer learning, which aims to transfer knowledge from an expert larger model (the `teacher') to a smaller one (the `student') without loss of fidelity. Generally, the student is trained to approach the functionalities of the teacher network by using the teacher network’s data.
For the production RL, accordingly, the aforementioned methods are appropriate to deal with the sim2real problem, the initial policy generated in the simulation environment (source domain) can be transferred to the actual environment, that is, the target domain.
\subsubsection{Continual Learning}
Continual learning is defined as the ability to continually learn over time by accommodating new knowledge while retaining previously learned experiences, which is also referred to as life-long learning. The fundamental element of continual learning is the knowledge base (KB) which contains all the knowledge accumulated in the past tasks. KB helps to train the new coming task, then KB will be updated accordingly. In other words, continual learning is a continuous learning process where the learner has performed a sequence of learning tasks. It leverages the information from the source domain(s) to help to learn in the target domain. The continual learning framework fits RL paradigms. There are mainly three categories of continual learning approaches (1) regularization, that is to avoid the model weights changing too much from the old weights; (2) replay, which means the data of the previous task and current task jointly train the current model; and (3) dynamic network architecture, which works by adding tuneable weights based on the current frozen model for learning more tasks \cite{liu2021lifelong}.
\subsubsection{Unsupervised domain transfer}
This technique aims to learn a map between sim and real so that RL agents trained in simulations can be adjusted for the expected data shift of the ``real''. It is an unsupervised technique in the sense that this map is learned using ``unpaired'' examples from the two-state spaces. A recent example is the cycle generative adversarial networks (CycleGAN) domain adaptation. CycleGAN has been initially proposed to couple the image-to-image translation tasks \cite{zhu2017unpaired}. It aims to learn a mapping between two distributions through a pair of GANs. One is used to adapt from the source to the target domain, and the other is in another way around. The cycle consistency loss is introduced to ensure that GANs could reproduce the original images continuously. With the deepening of CycleGAN research, lots of CycleGAN variants appear, like conditional CycleGAN, time-series CycleGAN, and RL-CycleGAN. They basically conform to CycGAN's pairwise training logic and are accompanied by specific loss function reconstructions. These architectures are pretty attractive to a sim2real problem, that is, sim2real is the source to target domain while real2sim is the target to the source domain. By CycleGAN, the mapping from sim to real is expected to be learned \cite{rao2020rl}.
\subsection{Generic Model Learning}
\subsubsection{Meta Learning}
Meta learning is the paradigm of learning to learn. It aims to learn a meta model with agile adaptive ability from a set of correlated learning tasks. The optimization direction of meta model follows the hidden distribution of tasks \cite{finn2017model}. So, this model can be adopted in a new scenario by zero or few shots learning. Meta learning is promising in solving sim2real. With DT's parallel validation and flexible configuration features, it is entirely operational to simultaneously generate a series of correlated tasks to train a meta model.
\subsubsection{Domain Randomization}
Domain randomization is similar to the learning frameworks mentioned above, while they are more close to the training skills of improving the generalisation ability of RL models. Domain randomization attempts to approximate or cover the distribution of the real-world environment by introducing randomness into the simulated environment, then the trained agent hopefully matches the real environment. For instance, in a visual robot control task, usually, the vision parameters in the training process could be randomised. Meanwhile, in the training process, introducing disturbances in environmental observation or reward acquiring is also a feasible method for a robust agent \cite{zhao2020sim}.
\section{Further Discussion of RL applications}
In the above sections, we discussed RL related sim2real issues, case studies and some learning-based solutions. In this section, we stream the scope down to RL applications. The potential concerns of RL applications that may exist in real networks and mechanisms for related algorithm development will be briefly discussed as part of the macro sim2real issue.

\subsection{Data Issues}
First, for network operators, to develop RL-based xApps or rApps for O-RAN, although DT can partially play the role of external environment and data source in the training stage. But in field validation, interaction with real networks is always unavoidable. Hence, data from multiple sources need to be collected, validated, enriched, transformed and stored in an integrated data pool, to meet the ongoing data requirements at each stage of validation and deployment or even re-training. That needs to be processed by data engineering processes, such as application of business rules, creation of KPIs, feature engineering, linkage of data tables according to network topology mapping, etc., which ultimately enables the RL applications according to the targeted use cases. O-RAN network is built on top of other system components such as IP networks and IT/Cloud infrastructures. The operation and maintenance of these systems are crucial for the overall network performance. It should be integrated into a holistic network management process that addresses all the components. Meanwhile, safety measures should be in place for interactions between RL applications and live networks.

\subsection{Environment Issues}
Second, there are two noteworthy environment-related issues, which cause a considerable impact on the effectiveness of RL applications. One is that the task reward can be delayed compared to the system state, which poses a critical challenge for the RL applications that interact with UEs. Due to the variation of wireless propagation channel and latency of network layer, The uncertain delay of getting UE reward feedback will cause chaotic training logic for agents. Another one is that the environment is partially observable. This problem is more obvious in telecommunications, where equipment-related information like the operating state, noise, efficiency etc. are fully not observable or stochastic for the agent. This also poses a challenge to the validity of the developed model \cite{dulac2020empirical}.

\subsection{Algorithm Design}
Finally, the design of the RL algorithm itself, which have been widely discussed in many papers. We list a couple of challenges that highly correlate with the next generation of cellular networks, which include but are not limited to:  \textit{(1) Safety concerns:} For RL models running on wireless networks, safety is important for service assurance as well as avoiding catastrophic performance decay. In the exploratory learning phase, potential safety restrictions that exist in the environments, agents, and actions should be considered and formalised in advance. The constrained MDP might be the solution \cite{russel2020robustCMDP}. \textit{(2) The complex state and high-dimensional discrete action space}: This problem is almost inevitable in real applications. For example, RL is used to control the transmission power of all cells in a region. That will lead to the failure of some RL algorithms and hard to generate a stable model. \textit{(3) Design of reward functions:} As the only model evaluation metric, the design of reward determines whether the trained model meets the training goal. For wireless RL applications, it is usually a  trade-off of the multi-objective reward functions. So an appropriate bonus setting for reward is crucial. \textit{(4) Reliability of RL models}: The RL model, due to its black-box prosperity, often raises concerns about its reliability. This is a problem that cannot be ignored for network operators who want to achieve risk-free.

\section{Conclusion}
The next generation of cellular networks is undoubtedly ML-driven. As a representative of them, O-RAN has a structure that naturally matches the ML paradigm and has great potential to promote the automation of future network infrastructures. Specifically, RL applications could bring more efficient and intelligent network operation strategies. 
But it is worth noting that few articles mention the sim2real problems encountered in the development of RL models in the context of O-RAN. 
This article intends to humbly summarize this issue to draw the attention of the research community. That is, it should be considered how to reliably transfer or deploy the RL model trained in the simulation environment to the real environment. In this regard, this article emphasises the main functionalities of high-fidelity DT in the process of the RL model development. We show the RL model deterioration caused by the UE distribution shift in the actual data and the advantages of DT-assisted RL model development, following with a series of sim2real-related learning methodologies. Finally, the article concludes with 
the analysis of the potential issues that existed in the realisation process of the RL applications in O-RAN.
We hope this work could motivate practical operations for RL driven applications in next generation networks. 

\section*{Acknowledgment}
This work was developed within the Innovate UK/CELTIC-NEXT European collaborative project on AIMM (AI-enabled Massive MIMO). This work has also been funded in part by the Next-Generation Converged Digital Infrastructure (NG-CDI) Project, supported by BT and Engineering and Physical Sciences Research Council (EPSRC), Grant ref. EP/R004935/1.

\bibliographystyle{IEEEtran} %
\bibliography{IEEEabrv,references} 

\section{biographies}

\begin{IEEEbiographynophoto}
  {Peizheng Li} received the B.Eng. degree in optoelectronic information engineering from Nanjing University of Posts and Telecommunications, China, in 2015, and the M.Sc. degree with distinction in image and video communications and signal processing from the University of Bristol, U.K., in 2019, where he continues pursuing the Ph.D. degree with the communication systems and networks group. His research interests include machine learning, radio localisation and radio access network.
\end{IEEEbiographynophoto}

\begin{IEEEbiographynophoto}
{Jonathan Thomas} received his M.Eng. Degree in Electronics and Communications Engineering from Cardiff University in 2016. He is now pursuing a Ph.D. degree within the communications systems and networks group at the University of Bristol. His research focuses on Multi-Agent Reinforcement Learning  communications networks.
\end{IEEEbiographynophoto}

\begin{IEEEbiographynophoto}
{Xiaoyang Wang} received her B.E. degree in electronic science and technology and her Ph.D. degree in signal and information processing from the University of Electronic Science and Technology of China (UESTC), Chengdu, China, in 2013 and 2018, respectively. She visited the University of Bristol, Bristol, UK, from 2017-2018. She has joined the Department of Electrical and Electronic Engineering, University of Bristol, as a Research Associate since 2018. She has worked on a variety of computer vision topics such as target detection, remote sensing image processing and visual tracking. Her current research focuses on machine learning, especially reinforcement learning, and its applications in next-generation network management.
\end{IEEEbiographynophoto}

\begin{IEEEbiographynophoto}
{Hakan Erdol}
received his B.S. and M.Sc Degrees in Electrical and Electronics Engineering from Karadeniz Technical University, Turkey., in 2013 and 2016 respectively. After working as a Research assistant at Karadeniz Technical university. He has continued at the University of Bristol pursuing a Ph.D. in Electrical and Electronics Engineering with a focus on Distributed Learning algorithms since 2020. His current research area revolves around Federated learning framework and Transfer learning techniques.
\end{IEEEbiographynophoto}

\begin{IEEEbiographynophoto}
{Abdelrahim Ahmad} is the Data Scientist Lead at Vilicom, where his most important work is related with the development of Data and Artificial Intelligence related applications focused on the Telecommunication industry. He has relevant and rich experience in AI, Data Architecture, Big Data ecosystems, and 5G and O-RAN technologies. Holder of two master's degrees in Big Data Systems \& Data Science and Web Sciences. Most remarkable experience have been achieved whilst leading data science projects, requiring the design and implementation of end-to-end big data analytical platforms to serve telecom businesses, he is mostly interested in building (near-RT, RT) platforms that combine O-RAN RAN Intelligent Controller with Reinforcement Learning applications.
\end{IEEEbiographynophoto}

\begin{IEEEbiographynophoto}
{Rui Inacio} is currently Vilicom’s Chief Technology Officer being responsible for the digitisation of the company’s business models mainly through the development of digital platforms that combine O-RAN technology, Big Data technologies and AI technologies. He has built a diversified career in the Mobile Communications industry, with a rich experience across multiple areas of expertise such as network management, network operations, network deployment and network support systems. He has both experience as an engineer and as a technologist that thinks of and about the network life-cycle and also thinks how to develop technology to assist engineers in this life-cycle. He hold a master degree in Electronics and Telecommunications Engineering and a master degree in Business and Technology Management. Currently he is mostly interested in contributing for the development of cognitive networks that are highly automate and autonomic across the whole network management life-cycle.
\end{IEEEbiographynophoto}

\begin{IEEEbiographynophoto}
{Shipra Kapoor} received her PhD in Electronic Engineering from University of York, UK in 2019, with the subject of her thesis being ‘Learning and Reasoning Strategies for User Association in Ultra-dense Small Cell Vehicular Networks’. Since 2020 she has been a member of Self-Learning Networks team in Applied Research at BT, where she is now a Research Specialist. Her current research interests include application of artificial intelligence and machine learning techniques in next generation wireless networks to resource and topology management, connected and autonomous vehicle networks, O-RAN architecture and cognitive radio network. 
\end{IEEEbiographynophoto}

\begin{IEEEbiographynophoto}
{Arjun Parekh} leads a BT research team exploring the uses of Machine Learning \& AI for the automation of optimisation \& planning in converged networks.  He led the development of processes \& tools to drive the rollout of the UK's first 4G \& 5G networks, and has over 20 years experience in Radio Network Management, Optimisation and Automation. Arjun has an MPhys in Physics from the University of Oxford and is a Visiting Industrial Fellow at the University of Bristol.
\end{IEEEbiographynophoto}

\begin{IEEEbiographynophoto}
{Angela Doufexi} received the B.Sc. degree in physics from the University of Athens, Greece, in 1996; the M.Sc. degree in electronic engineering from Cardiff University, Cardiff, U.K., in 1998; and the Ph.D. degree from the University of Bristol, U.K., in 2002. She is currently a Professor of Wireless networks at the University of Bristol. Her research interests include vehicular communications; new waveforms, resource allocation, massive MIMO and multiple antenna systems, mmWave communications and sixth-generation communications systems. She is the author of over 200 journal and conference papers in these areas.
\end{IEEEbiographynophoto}

\begin{IEEEbiographynophoto}
{Arman Shojaeifard} has +9 years of post-PhD experience in research, development, and standardization of radio transmission technologies and architectures. He is currently an R\&I Senior Manager at InterDigital Europe where he serves as the Chair of the ETSI Industry Specification Group on Reconfigurable Intelligent Surfaces (ISG RIS) and the Technical Manager of the Innovate-UK/CELTIC-NEXT European collaborative R\&D project on AIMM (AI-enabled Massive MIMO). Prior to InterDigital, he was a Senior Researcher and 3GPP RAN1 WG Delegate at British Telecommunications plc. He received his PhD degree in Wireless Communications from King’s College London in 2013.
\end{IEEEbiographynophoto}

\begin{IEEEbiographynophoto}
{Robert J. Piechocki} is a Professor of Wireless Systems at the University of Bristol. His research expertise is in the areas of Connected Intelligent Systems, Wireless \& Self-Learning Networks, Information and Communication Theory, Statistics and AI. Rob has published over 200 papers in peer-reviewed international journals and conferences and holds 13 patents in these areas. He leads wireless connectivity and sensing research activities for the IRC SPHERE project (winner of 2016 World Technology Award). He is a PI/CI for several high-profile projects in networks and AI funded by the industry, EU, Innovate UK and EPSRC such as NG-CDI, AIMM, OPERA, FLOURISH, SYNERGIA. He regularly advises the industry and the Government on many aspects related to connected intelligent technologies and data sciences.
\end{IEEEbiographynophoto}

\end{document}